\documentclass{ecai}

\usepackage{latexsym}
\usepackage{amssymb}
\usepackage{amsmath}
\usepackage{amsthm}
\usepackage{booktabs}
\usepackage{enumitem}
\usepackage{graphicx}
\usepackage{color}

\usepackage{algorithm}
\usepackage[noend]{algorithmic}
\usepackage{multirow}
\usepackage[
nameinlink
]{cleveref}
\usepackage{svg}
\usepackage{xspace}
\usepackage{subcaption}

\newtheorem{theorem}{Theorem}

\newtheorem{proposition}[theorem]{Proposition}

\newtheorem{definition}{Definition}

\newcommand{\BibTeX}{B\kern-.05em{\sc i\kern-.025em b}\kern-.08em\TeX}

\newcommand{\algo}{\texttt{CCBNet}\xspace}
\newcommand{\augment}{\texttt{CABN}\xspace}
\newcommand{\infer}{\texttt{SAVE}\xspace}
\newcommand{\algoJ}{\texttt{CCBNetJ}\xspace}

\begin{document}

\begin{frontmatter}

\title{CCBNet: Confidential Collaborative Bayesian Networks Inference}

\author[A]{\fnms{Abele}~\snm{M\u{a}lan}\orcid{0009-0002-4493-7439}\thanks{Corresponding Author. Email: abele.malan@unine.ch.}}
\author[B]{\fnms{J\'er\'emie}~\snm{Decouchant}\orcid{0000-0001-9143-3984}}
\author[C]{\fnms{Thiago}~\snm{Guzella}\orcid{0000-0002-1374-7379}}
\author[A,B]{\fnms{Lydia}~\snm{Chen}\orcid{0000-0002-4228-6735}} 

\address[A]{University of Neuch\^atel}
\address[B]{Delft University of Tecnhology}
\address[C]{ASML}

\begin{abstract}
Effective large-scale process optimization in manufacturing industries requires close cooperation between different human expert parties who encode their knowledge of related domains as Bayesian network models. For instance, Bayesian networks for domains such as lithography equipment, processes, and auxiliary tools must be conjointly used to effectively identify process optimizations in the semiconductor industry. However, business confidentiality across domains hinders such collaboration, and encourages alternatives to centralized inference.
We propose \algo{}, the first \textbf{C}onfidentiality-preserving \textbf{C}ollaborative \textbf{B}ayesian \textbf{Net}work inference framework. \algo{} leverages secret sharing to securely perform analysis on the combined knowledge of party models by joining two novel subprotocols: (i) \augment{}, which augments probability distributions for features across parties by modeling them into secret shares of their normalized combination; and (ii) \infer{}, which aggregates party inference result shares through distributed variable elimination. We extensively evaluate \algo{} via 9 public Bayesian networks. Our results show that \algo{} achieves predictive quality that is similar to the ones of centralized methods while preserving model confidentiality. We further demonstrate that \algo{} scales to challenging manufacturing use cases that involve 16--128 parties in large networks of 223--1003 features, and decreases, on average, computational overhead by 23\%, while communicating 71k values per request. Finally, we showcase possible attacks and mitigations for partially reconstructing party networks in the two subprotocols.
\end{abstract}

\end{frontmatter}

\section{Introduction}

\begin{figure*}[tb]
\centering
\includegraphics[width=\linewidth]{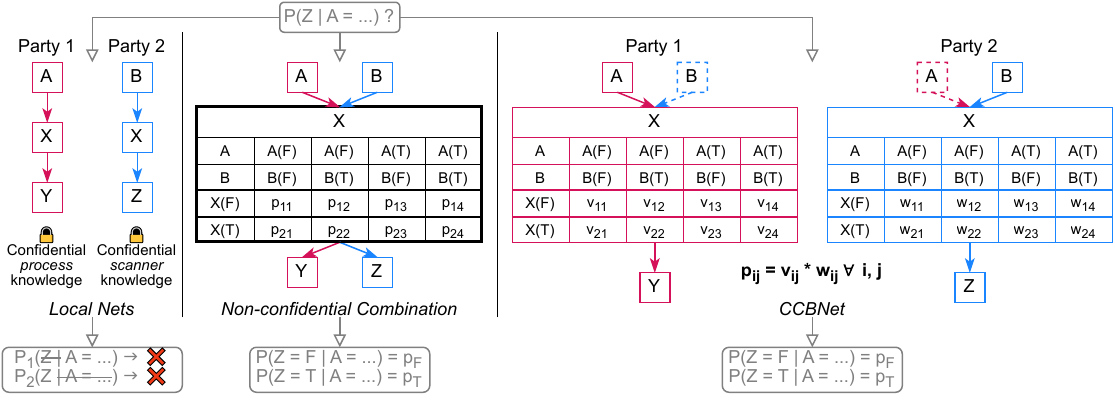}
\caption{\textbf{Sample Bayesian networks collaboration in semiconductor manufacturing:} Parties' separate models overlap within feature $X$. Party 1 features $A$ and $Y$ inform about the etching process and minimum feature size. Party 2 features $B$ and $Z$ inform about the wafer table and scanner overlay characteristics. After observing $A$, inferring updated state probabilities for $Z$ requires propagating information between $A$, $B$, and $X$, which is impossible by merely averaging model outputs. The typical non-confidential approach reveals the combined graph and probability tables to parties. \algo{} exchanges minimal structural information amongst parties, and secret shares overlap feature tables, reconstructing the final result by distributed inference.}
\label{fig:comp}
\end{figure*}

Improving productivity and quality standards in manufacturing demands effectively expressing complex interactions between domain items. Bayesian networks (BNs) are commonly adopted to graphically model causality in manufacturing~\cite{manufacture_nannapaneni}, with nodes representing features and directed edges showing dependencies. An essential trait of these models is their ability to specify arbitrary input and output features for each query instead of having them fixed.

Let us consider the use case of semiconductor manufacturing. Pursuing ever smaller size chips at a high yield~\cite{semiconductor_yang} entails cooperation between many specialized parties that must protect their trade secrets. More specifically, fab operators encode their insight into settings dictating the production process in a BN. Similarly, vendors of equipment like scanners, construct BNs that describe the inner workings of their machines. \Cref{fig:comp} illustrates such a scenario. Pooling together parties' knowledge allows higher-quality analysis that optimize production environments, leading to new business opportunities. Simultaneously, the need to protect intellectual property expressed within party models calls for confidential collaboration.

Existing studies on collaborative inference for BNs \textbf{disregard model confidentiality constraints} or make concessions about which party information can be merged and how. Many have focused on centralized scenarios that combine local BNs' knowledge into a larger one~\cite{combine_delsagrado,combine_feng} without protecting confidential knowledge within the input networks and global output. Models get stitched together based on common nodes, and remain in the final representation as largely unaltered sub-models whose encoded knowledge is easily inspectable. \citet{distr_pavlin} proposes a distributed combination variant that partially preserves confidentiality by maintaining the locality of combined party models. However, it leaks information between parties when connecting them and does not allow merging inner graph nodes having both parents and children. \citet{dist_tedesco} maintains the confidentiality of how nodes are linked within parties but only allows propagating information between them in a fixed sequential order. \cite{bcc_kim} is another distributed approach with similar confidentiality properties but even greater compatibility restrictions by requiring models to have identical inputs and outputs.

In this paper, we propose \algo{}, the first confidential, collaborative BNs inference framework that combines knowledge of multiple parties involved in inference queries through a novel secret sharing scheme. \algo{} does not require a trusted third party, and protects confidentiality at both the levels of party models and data instances. The two key components of \algo{} are: (i) confidential sharing of a normalized combination of features' probability distributions across all overlapping parties; and (ii) distributed inference based on variable elimination for aggregating party results. The novelty of the augmentation procedure lies in constructing discrete conditional probability distributions for all features present in more than one party. These features represent secret shares of a combined and normalized distribution from a centralized scenario without exposing any party's initial probability function. To evaluate \algo{}, we simulate knowledge compartmentalization over different starting public BNs. Moreover, we demonstrate possible attacks against the framework, their limitations, and ways to protect against them.

In summary, we make the following contributions:
\begin{itemize}
    \item We devise a confidential, collaborative framework for BNs, \algo{}, that satisfies the needs of industry use cases like process optimization in manufacturing.
    \item We design a novel secret sharing-based protocol, \augment{}, to confidentially augment the conditional probability function of overlapping features in parties.
    \item We define \infer{}, a new method for performing distributed inference on augmented party models backed by variable elimination that aggregates their result shares.
    \item We evaluate \algo{} over various scenarios based on nine public BNs. Our results indicate that \algo{}'s predictive accuracy is similar to those of non-confidential centralized alternatives and, for many collaborators in large networks, that \algo{} decreases the computational overhead by 23\% on average when 71k values are communicated per request.
    \item We discuss attacks on the two \algo{} subprotocols, possible mitigation strategies and their trade-offs.
    
\end{itemize}

\section{Background \& Related Studies}

\subsection{Background}

We review relevant BN notion with \Cref{fig:comp} as reference.

\textbf{Bayesian Networks} are probabilistic graphical models that maintain explicit conditional probability distributions (CPDs) for features whose dependencies form a directed acyclic graph~\cite{bn_stephenson}. Within such context, features are often referred to as (probabilistic) variables or (graph) nodes. Shown in \Cref{fig:comp}, features and the influences between them give the graph nodes and edges, respectively. Principally for performance and interpretability, features in practical applications are generally discrete, with CPDs specifically embodying conditional probability tables. Learning may be driven by data, human experts, or both~\cite{bn_koller, learning_daly}, as with other human-readable models like decision trees. Automated learning discovers the graph structure and then populates CPD parameters from training data. Manual learning is desirable when incorporating concepts with known governing rules that need no approximation from observations. Examples are physical phenomena or, in manufacturing, human-engineered processes and tools, like Party 2's scanner wafer table from \Cref{fig:comp}.

\textbf{Inference in BNs} finds updated posterior probabilities for the states of target variables, given the observed states of any evidence variables~\cite{bn_russel,causality_pearl}. As general inference is NP-Hard, approximate algorithms help decrease computation costs compared to exact ones while sacrificing some precision in the result. The main exact inference techniques are variable elimination and junction tree belief propagation, which decomposes the network into a tree of variable clusters, running variable elimination within them and then disseminating updates between neighbors by message-passing~\cite{bn_koller}. In \Cref{fig:comp}'s query, $Z$ is the target, given some observed state of $A$.

\textbf{Computation in discrete BNs} relies on a few base operations for propagating information: normalization, reduction, marginalization, and products~\cite{bn_koller}. We outline them with help from the non-confidential combination in \Cref{fig:comp}. Normalizing a CPD divides its entries by their column sum. Thus column summations, like $p_{11} + p_{21}$, would equal $1$. Reduction and marginalization remove variables from a CPD by fixing their states or, respectively, summing them out. Reducing or marginalizing $A$ from $X$ leaves $B$ as its sole parent. Flattening CPD structures yields factors that specify a value for each state combination of their variables without discriminating between the child and parents. Previous operations apply to both representations. Products operate on CPDs of the same variable or factors and create a new CPD/factor over the input variables' union, where each entry is the multiplication of the corresponding ones in the original representations. The product of $X$ and $Z$'s factors, thus, additionally contains $A$ and $B$.

\textbf{Markov random fields (MRFs)} are a sibbling model of BNs, backed by undirected graphs, into which every BN is easily transformable via moralization~\cite{mrf_li, bn_scutari}. Apart from lacking acyclicity constraints, MRFs directly define parameters as factors and can deal with scenarios where edge directionality is unspecified, but BNs are more compact and efficient for generative use. For inference, BN properties and algorithms remain applicable. Thus, when aciclicity constraints are unsatisfiable, MRFs can still perform probabilistic inference like in BNs.

\subsection{Prior Art}

We identify two high-level categories of collaborative analysis for BNs: single- and multi-model. The first creates one global model, while the other keeps local models separate, merging their results.

\textbf{Single-model} approaches harness party data instances or models but neglect confidentiality. Federated learning discovers models~\cite{flbn_ng,flbn_gao,flbn_huang,flbn_abyaneh,flbn_vandaalen} from private party instances with a coordinator, but fully decentralized alternatives exist~\cite{approximate_campbell,approximate_gholami}. Direct network combination fuses structure~\cite{combine_delsagrado,combine_alrajeh} and parameters~\cite{combine_feng} from party models.

\textbf{Multi-model} methods have parties work together during inference to produce a complete analysis result. \citet{dist_tedesco} chains model without exchanging their contents but only allows using them one at a time in a predetermined order. Less confidential but more flexible, \citet{distr_pavlin} fuses party networks based on common nodes but still requires them to be roots or leaves in the party's directed acyclic graph. \citet{collaborative_ypma} patents a collaborative solution for industrial processes, only mentioning data confidentiality preservation by anonymization. \citet{bcc_kim} runs models autonomously and only averages their final outputs, maintaining confidentiality but expecting models to share inputs and outputs.

In summary, single-model techniques break confidentiality by centralizing knowledge, and multi-model ones trade modeling power for it. \algo{} addresses both concerns.

\section{\algo{}}

\begin{figure*}[tb]
\centering
\begin{subfigure}[b]{0.7475\textwidth}
    \centering
    \includegraphics[width=\textwidth]{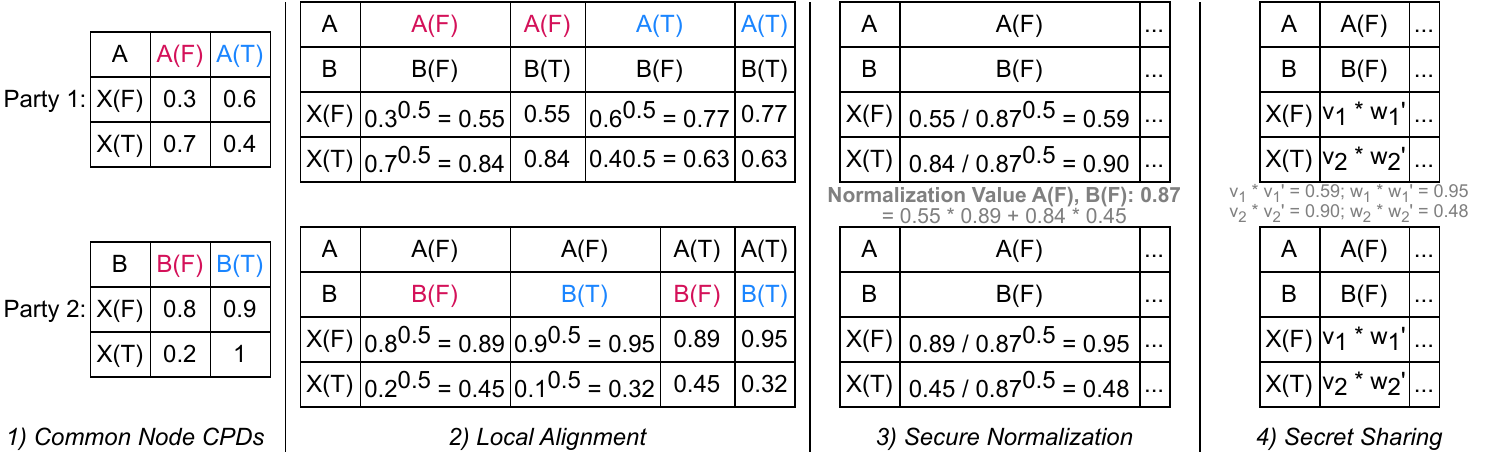}
    \caption{Key Steps of computing augmented CPDs}
    \label{subfig:augment}
\end{subfigure}
\hfill
\begin{subfigure}[b]{0.2475\textwidth}
    \centering
    \includegraphics[width=\textwidth]{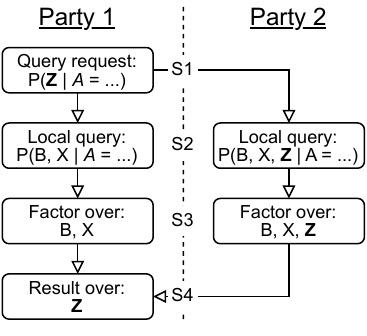}
    \caption{Distributed inference flow}
    \label{subfig:infer}
\end{subfigure}
\hfill
\vfill
\caption{\augment{} \& \infer{} steps for \Cref{fig:comp} scenario}
\label{fig:algo}
\end{figure*}

We propose a framework, \algo{}, for secure distributed analysis over a related set of confidential and discrete BNs. \algo{} is composed of two key steps, (i) \augment{} augments the BNs of parties through overlapping variables, and (ii) \infer{} performs joint inference on them.

\textbf{The assumptions} we make are that features from different parties have the same name only if they represent the same concept, and the independence, across parties, of distinct parents for the same node reasonably approximates the ground truth. These are shared by previous BN combination works. Thus, names identify the overlapping (common) nodes between models, giving the contact points for graph fusion. Since features modeled by parties may be any subset of those from the full domain, modeling direct interactions between their non-overlapping variables requires great amounts of often unavailable information.

\textbf{Our adversarial model} includes semi-honest parties that follow the protocol while trying to abuse gained information~\cite{sec_goldreich} but do not collude. Further, no trusted third party exists. The goal is to protect all network parameters and only disclose common nodes' structure/state information amongst parties containing them.

\subsection{Confidentially Augmented Bayesian Networks}

\begin{algorithm}[tb]
\begin{algorithmic}[1]
    \FOR{pX, pY $\leftarrow$ Parties $\times$ Parties} \label{aug-l1}
        \FOR{\it{node} $\leftarrow$ \textit{PrivateNodeIntersect}(pX, pY)} \label{aug-l2}
            \STATE overlaps[node] $\cup\leftarrow$ \{pX, pY\} \label{aug-l3}
        \ENDFOR
    \ENDFOR
    \FOR{node, parties $\leftarrow$ overlaps} \label{aug-l4}
        \STATE states $\leftarrow \bigcup_{p \in parties}$ \textit{ObfuscatedStatesCPD}(p, node) \label{aug-l5}
        \STATE idCPD $\leftarrow$ \textit{IdentityCPD}(node, states) \label{aug-l6}
        \STATE weightSum $\leftarrow \sum_{p \in parties}$ p.weight \label{aug-l7}
        \FOR{p $\leftarrow$ parties} \label{aug-l8}
            \STATE p.CPD[node]  $*\leftarrow$ idCPD \label{aug-l9}
            \STATE p.CPD[node] $**\leftarrow$ p.weight / weightSum \label{aug-l10}
        \ENDFOR
        \FOR{col $\leftarrow$ idCPD.cols} \label{aug-l11}
            \STATE colsHE $\leftarrow \bigcup_{p \in parties}$ \textit{HE}(p.CPD[node $|$ col]) \label{aug-l12}
            \STATE normVal $\leftarrow$ \textit{L1HadamardProdHE}(colsHE) \label{aug-l13}
            \FOR{p $\leftarrow$ parties} \label{aug-l14}
                \STATE p.CPD[node $|$ col] $/\leftarrow$ normVal$^{1 / |parties|}$ \label{aug-l15}
            \ENDFOR
        \ENDFOR
        \STATE \textit{MultipicSecretShare}($\bigcup_{p \in parties}$ p.CPD[node]) \label{aug-l16}
    \ENDFOR
\end{algorithmic}
\caption{\augment{}}
\label{alg:aug}
\end{algorithm}

We now present the \augment{}\footnote{\textbf{C}onfidentially \textbf{A}ugmented \textbf{B}ayesian \textbf{N}etworks} protocol, which updates local CPDs for overlap variables to hold secret shares of their normalized central combination while protecting the initial probabilities. \Cref{alg:aug} details the four steps of the protocol, illustrated in \Cref{subfig:augment}: (i) private common node identification; (ii) local alignment; (iii) secure normalization; and (iv) secret sharing. The protocol updates parties when the number of changes in local networks passes a set threshold.

\textbf{Overview.} Structurally, \augment{} imitates a union of the involved networks of \citet{combine_delsagrado}, and parameter-wise, it follows \citet{combine_feng} but replaces the superposition operator with the geometric mean. We use the union instead of the more complex ruleset of \citet{combine_feng} for deciding which overlapping node parents to retain.  Such a more straightforward logic reduces the surface area for attacks and allows for combining more than two BNs at a time, lowering the number of communication rounds. For fusing probabilities, the geometric mean enables a multiplication-based secret sharing scheme in \augment{} where reconstruction happens automatically during distributed inference when computing intermediary party factor products. The mean also outperforms the superposition in our centralized tests. Having described the general strategy, we continue with the protocol phases.

\textbf{Step 1: Private Common Node Identification.} \augment{} starts with party pairs identifying their common nodes like in the central case, albeit privately. We use private set intersection protocol~\cite{psi_decristofaro} to attain confidentiality. The pseudocode highlights this step in ll. \ref{aug-l1}-\ref{aug-l3}. Only parties that have updates need to recalculate their intersection with the others. Outside the private intersection context, node and state names are communicated obfuscated to prevent information leakage about which nodes are modeled by which party. Parties can choose any unique representation for non-overlapping nodes, but involved parties agree on an obfuscated representation for overlapping ones.

\textbf{Step 2: Local Alignment.} After parties know which local nodes overlap with which peers, they start solving overlaps by exchanging composition and weight information about their local CPDs and independently updating local representations accordingly (lines \ref{aug-l4}-\ref{aug-l10}). First, the obfuscated union of CPD nodes and states for overlapping CPDs is determined (line \ref{aug-l5}). From it, an identity CPD containing all the parents across parties gets created for the union (line \ref{aug-l6}). An identity CPD (or factor) has all entries equal to 1, so its product with another replicates the later's columns over their joint state space. The initial CPD gets replaced by the product with the identity in each party, giving all overlap CPDs the same shape (line \ref{aug-l9}), as seen in \Cref{subfig:augment}. Parties have a public weight representing confidence in their BN, which in the default unweighted case is 1. They compute the sum of their weights (line \ref{aug-l7}) and individually raise the entries of their CPD to the ratio between their weight and the sum, computing the partial geometric mean (line \ref{aug-l10}). By the exponent product rule $(XY)^k = X^kY^k$, the CPDs' product already yields the unnormalized central combination CPD.

\textbf{Model Weighting.} As previously mentioned, \augment{} allows weighting CPDs through the geometric mean, contrary to previous BN works that cover parameter fusion. A natural integration of unequal weighting of inputs is another advantage of using a geometric mean instead of the superposition from \citet{combine_feng}. We implement weights at the model level as 0-1 values, encoding the human expert's confidence in the network or data availability for algorithmic learning. Nevertheless, weighting can be applied at the CPD level.

\textbf{Step 3: Secure Normalization.} Homomorphic encryption (HE)~\cite{ckks_cheon} enables privately computing column normalization values (lines \ref{aug-l11}-\ref{aug-l15}). One party is elected to generate the public/private key pair, while another does the computation. All parties receive the public key and send their encrypted columns (line \ref{aug-l12}) to the party that calculates normalization value ciphers (line \ref{aug-l13}), which the private key party later receives, decrypts, and shares with the rest. A column normalization value is the sum of entries obtained by multiplying corresponding party columns element-wise. Letting $P_{ij}$ denote party $i$ CPD column $j$, the calculated value is $||\odot_i P_{ij}||_1$. Then, the $K$ overlapping parties individually divide each column by the $K$-th root of the appropriate normalization value (line \ref{aug-l14}), so their factor product is the normalized geometric mean. \Cref{subfig:augment} shows an example.

Because local columns no longer sum to 1 after exponentiation, even in a two-party overlap where the variable has only two possible states, a party cannot reconstruct the other's entries by only knowing the normalization values and its own entries. Furthermore, vital for HE schemes in practice, we know that the number of consecutive multiplications needed for each column is equal to the party count, which allows configuring the scheme accordingly. Functional encryption, in which completing the desired computation also decrypts the output~\cite{fe_boneh}, can be a more viable choice, but existing implementations have overly stringent limits on the number of inputs and complexity of the applied functions. Using a secret sharing scheme (SSS)~\cite{smpc_cramer} instead of HE is also possible, but we favor decreasing the communication count over computing overhead for this step. An SSS has the advantage of requiring fewer computational resources and being more robust against collusion. However, it requires communication for each multiplication operation and, depending on the scheme, the presence of a third party. Despite the expectation that \augment{} needs to run more rarely than inference, we still favor optimizing for message count, as high communication latency is likelier to be a bottleneck than processing for envisioned deployments.

\textbf{Step 4: Secret Sharing.} Finally, to combat party parameter leaks at inference, we secret share~\cite{fairness_kilbertus} the CPD entries of parties in each overlap through a multiplication-based scheme (line \ref{aug-l16}). The scheme allows performing the specific operations needed for inference while exchanging a similar number of messages as HE and achieving much better scaling in terms of computation. The common shape of updated local CPDs facilitates the procedure. In the classic additive secret sharing scheme, a secret value is split into shares distributed amongst parties whose sum is the secret. It allows efficient and secure computation of expressions summing multiple secret values and applying other operations involving non-secret values. Parties perform the computation with their local share of each secret and all aggregate their results to reconstruct the answer. The utilized scheme functions similarly but uses multiplication as the base operation instead. Reconstruction happens during inference as before, with no extra overhead since party CPDs contain different information that needs merging regardless. The appendix provides more information on the procedure. \Cref{subfig:augment} exemplifies the share splitting.

\emph{Handling potential cycles.} To avoid compatibility restrictions between combinable BNs, if solving overlaps creates a cycle, the distributed global network gets treated as an MRF, with no changes to the inference, which operates on factors regardless. Edges that form cycles in the BN are effectively incorporated into the moralized MRF and treated as undirected. Since the main target is to not share the complete combined network, the readability advantages of BNs are not a concern in the joint global model. Deciding which edge to remove from a cycle often requires unavailable information and threatens confidentiality. Alternatives like treating all nodes within a cycle as a single node~\cite{collaborative_ypma} are coarse-grained and threaten confidentiality.

\subsection{\infer{}: Share Aggregation Variable Elimination}

\begin{algorithm}[tb]
\textbf{Input}: Q=\{x, y, $\dots$\}, E=\{a\textsubscript{i}, b\textsubscript{j}, $\dots$\}\\
\textbf{Output}: Factor
\begin{algorithmic}[1]
    \STATE auxFacts $\leftarrow$ \{\} \label{inf-l1}
    \FOR{party $\leftarrow$ Parties} \label{inf-l2}
        \STATE partyFacts $\leftarrow \bigcup_{cpd \in party.CPDs}$ Factor(cpd) \label{inf-l3}
        \STATE partyQ $\leftarrow$ Q $\bigcup$  \textit{OverlapNodesAndParents}(party) \label{inf-l4}
        \STATE auxFacts $\cup\leftarrow$ \textit{VarElim}(partyQ, E, partyFacts, True) \label{inf-l5}
    \ENDFOR
    \RETURN \textit{VarElim}(Q, \{\}, auxFacts, False) \label{inf-l6}
\end{algorithmic}
\caption{\infer{}}
\label{alg:inf}
\end{algorithm}

\infer{}\footnote{\textbf{S}hare \textbf{A}ggregation \textbf{V}ariable \textbf{E}limination} is the inference protocol that has all parties run variable elimination locally before aggregating their outputs into the final factor. \Cref{alg:inf} describes its steps, and \Cref{subfig:infer} visualizes them for an example query. The main variable elimination algorithm loop yields a set of leftover factors that share no dependencies and whose normalized product represents the eventual output. In the pseudocode, \texttt{VarElim} returns the result of the complete algorithm when the final argument flag is not set but stops early, returning the set of leftover factors otherwise. The party requesting inference sends the evidence and target variables to all others in obfuscated form (S1 in fig.). Parties run the query locally, adding to the target set their overlapping variables and direct parents (S2 in fig.). Unmodeled variables are ignored. Their intermediate factors, representing shares of the final result, are sent back to the requesting party in obfuscated form (S3 in fig.), which runs a final round of variable elimination for the result after receiving all replies (S4 in fig.). Adding overlap variables and parents to local target sets avoids marginalization before reuniting all information related to them. Marginalization entails summing values, which cannot happen locally with the chosen SSS, so we delay it until implicit share reconstruction within the last variable elimination call at the initiating party.

\textbf{Queryable Nodes.} To maintain confidentiality, parties can only specify modeled variables in inference by default, even if the result still reflects the effect of prior knowledge about others, so we propose mechanisms for expanding the set of possible queries. The first involves all parties that own a node agreeing to expose its unobfuscated name and states with select others to use as a target or evidence. Doing so only requires revealing a node's existence, not its place in the network(s). The other mechanism implicitly enhances evidence with the help of some key shared between parties (e.g., timestamp, product batch identifier). If a query request also includes a value for the shared key, parties incorporate any observations for the key's value as evidence during their local inference step. Parties do not have to disclose the value of the observed data, but the query output is  the same as if it had been part of the initial evidence.

\textbf{Communication Properties.} The number of messages exchanged within an inference request is of magnitude $O(N)$ (where $N$ is the number of parties), but the size of the messages varies. Regarding count, the requester sends out $N-1$ messages and receives the same amount of replies adding up to $2N-2$. The size of the messages, particularly replies, varies greatly depending on the number and complexity of the responder's overlaps and the query itself.

\section{Performance Evaluation}

The following section shows \algo{} achieving the predictive ability of centralized non-confidential solutions in a confidential setting, with reasonable computation time and communication size. The appendix includes further details.

\begin{table}[tb]
\tabcolsep=0.1cm
\centering
\scalebox{0.925}{
\begin{tabular}{ccccc}
\hline
Class & Name & \#Nodes & \#Edges & \#Params \\ \hline
\begin{tabular}[c]{@{}c@{}}Small\\ ($<$ 20 Nodes)\end{tabular} & ASIA & 8 & 8 & 18 \\ \hline
\multirow{3}{*}{\begin{tabular}[c]{@{}c@{}}Medium\\ (20-49 Nodes)\end{tabular}} & CHILD & 20 & 25 & 230 \\
 & ALARM & 37 & 46 & 509 \\
 & INSURANCE & 27 & 52 & 1008 \\ \hline
\begin{tabular}[c]{@{}c@{}}Large\\ (50-99 Nodes)\end{tabular} & WIN95PTS & 76 & 112 & 574 \\ \hline
\multirow{3}{*}{\begin{tabular}[c]{@{}c@{}}Very Large\\ (101-999 Nodes)\end{tabular}} & ANDES & 223 & 338 & 1157 \\
 & PIGS & 441 & 592 & 5618 \\
 & LINK & 724 & 1125 & 14211 \\ \hline
\begin{tabular}[c]{@{}c@{}}Massive\\ ($\geq$ 1000 Nodes)\end{tabular} & MUNIN2 & 1003 & 1244 & 69431
\end{tabular}
}
\caption{Evaluation datasets w/ node, edge, parameter counts}
\label{tab:net-stats}
\end{table}

\begin{figure*}[tb]
\centering
\begin{subfigure}[b]{0.33\textwidth}
    \centering
    \includegraphics[width=\textwidth]{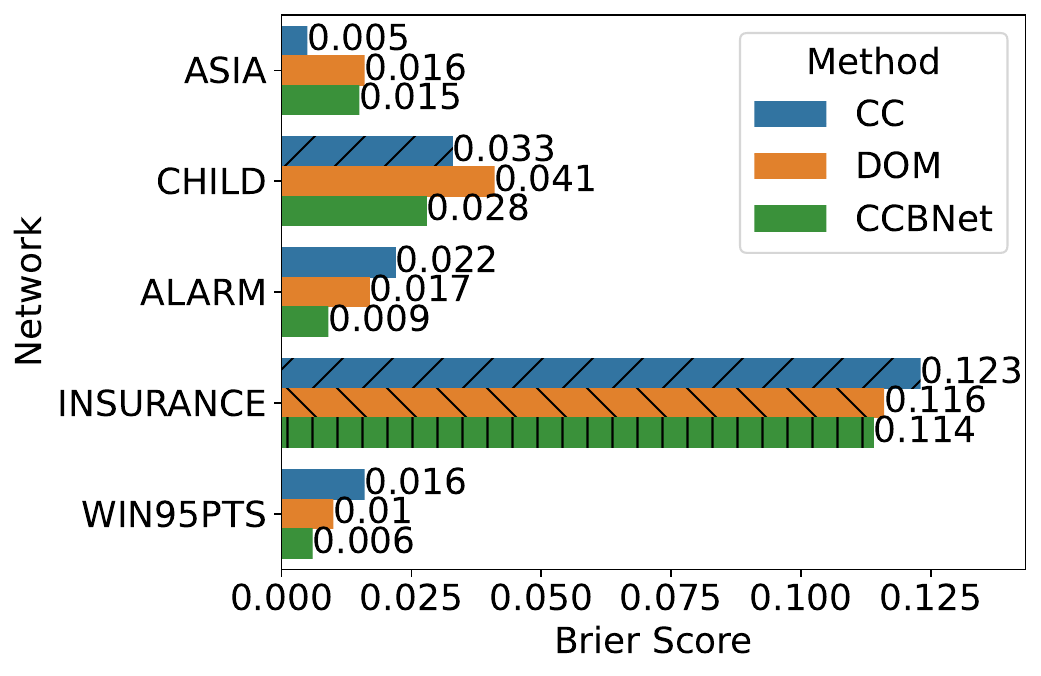}
    \caption{Brier scores relative to original network, related splits}
    \label{subfig:brier-related}
\end{subfigure}
\hfill
\begin{subfigure}[b]{0.33\textwidth}
    \centering
    \includegraphics[width=\textwidth]{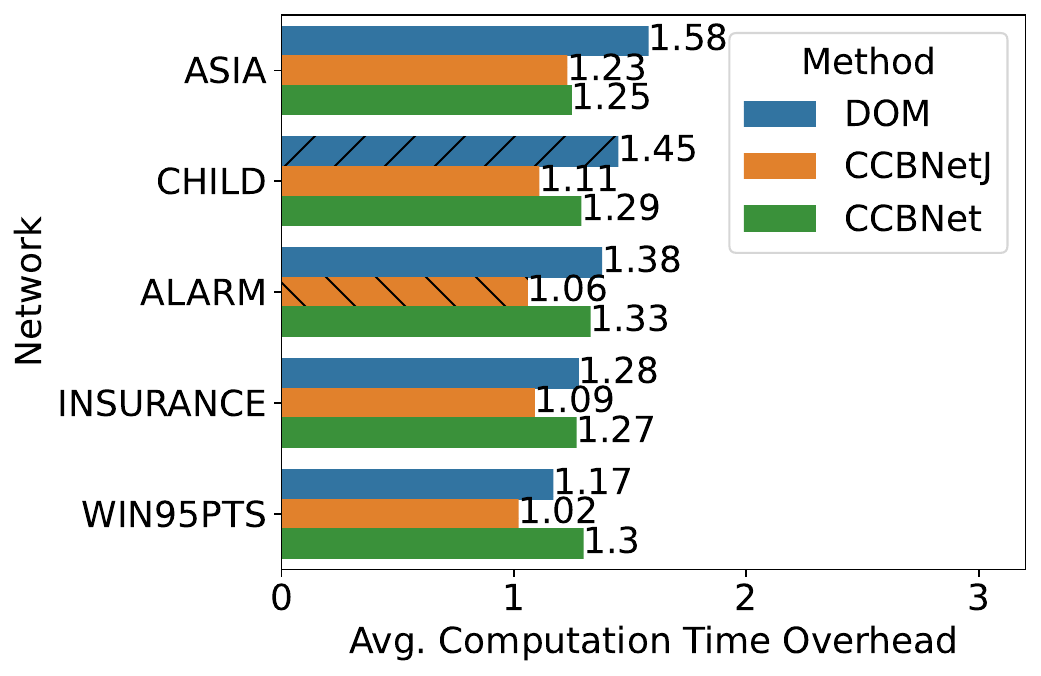}
    \caption{Average computation time overhead relative to original network, related splits}
    \label{subfig:time-related}
\end{subfigure}
\hfill
\begin{subfigure}[b]{0.33\textwidth}
    \centering
    \includegraphics[width=\textwidth]{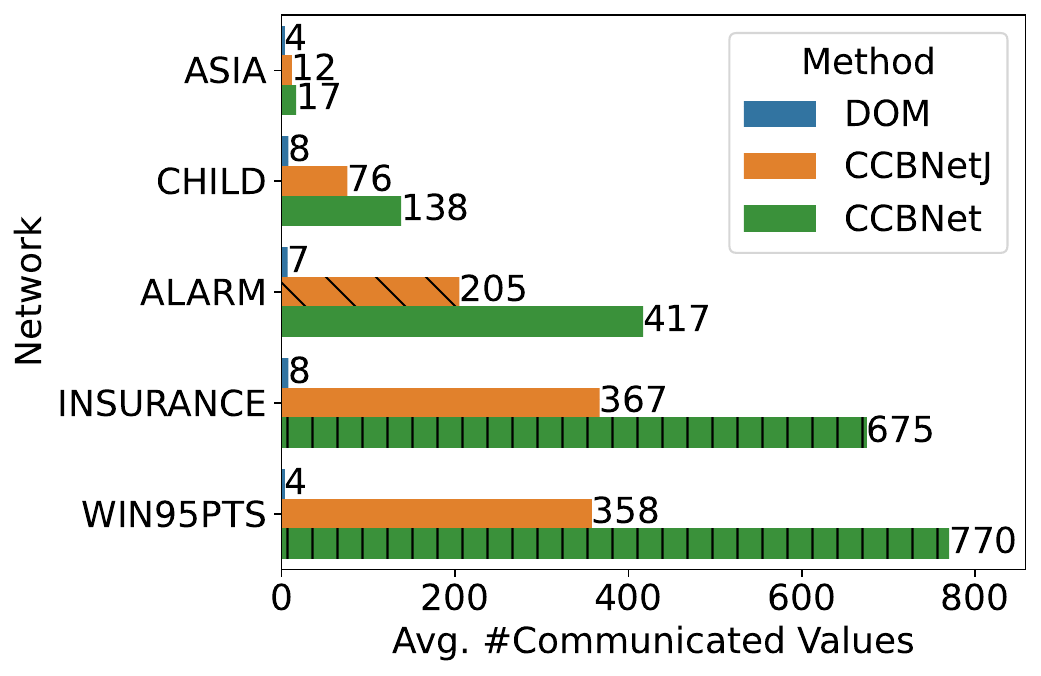}
    \caption{Average \#communicated values rounded to nearest integer, related splits}
    \label{subfig:comm-related}
\end{subfigure}
\hfill
\vfill
\begin{subfigure}[b]{0.33\textwidth}
    \centering
    \includegraphics[width=\textwidth]{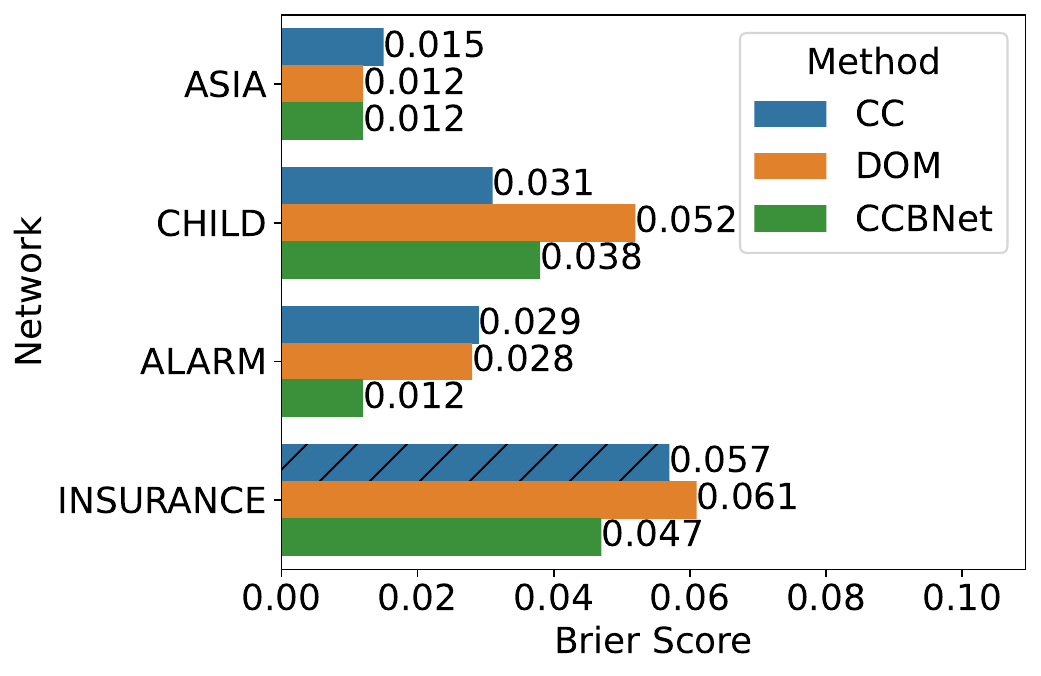}
    \caption{Brier scores relative to original network, random splits}
    \label{subfig:brier-random}
\end{subfigure}
\hfill
\begin{subfigure}[b]{0.33\textwidth}
    \centering
    \includegraphics[width=\textwidth]{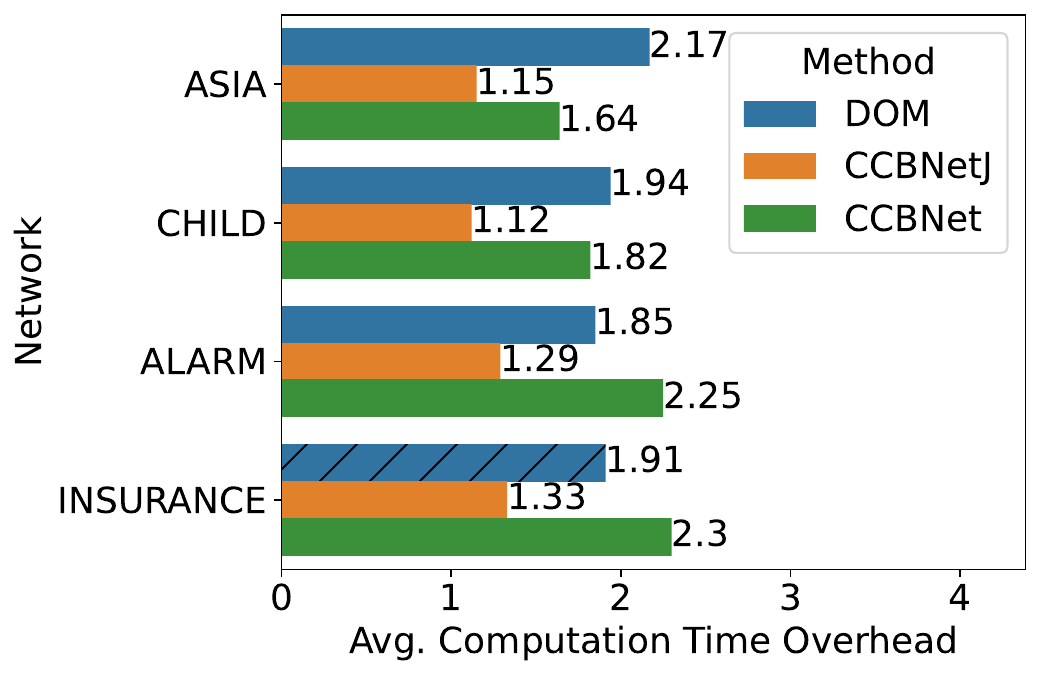}
    \caption{Average computation time overhead relative to original network, random splits}
    \label{subfig:time-random}
\end{subfigure}
\hfill
\begin{subfigure}[b]{0.33\textwidth}
    \centering
    \includegraphics[width=\textwidth]{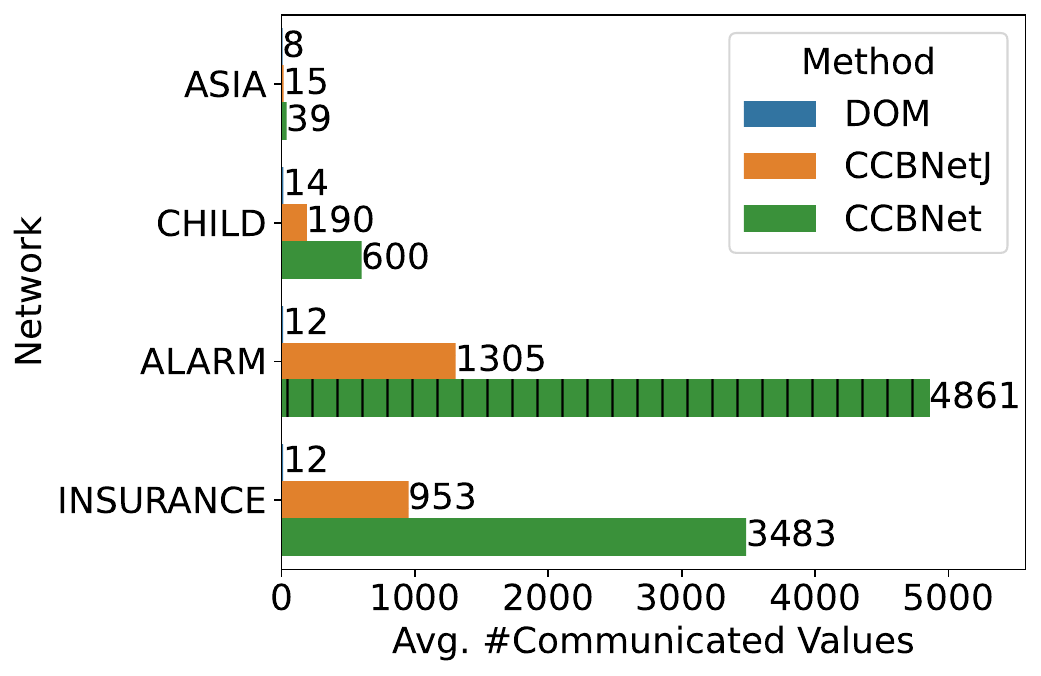}
    \caption{Average \#communicated values rounded to nearest integer, random splits}
    \label{subfig:comm-random}
\end{subfigure}
\hfill
\vfill
\caption{Results for 4 parties, 30\% of vars in \textgreater{}1 party (lower is better for all)}
\label{fig:res}
\end{figure*}

\subsection{Setup}

Our experiments evaluate average predictive performance, computation overhead, and communication cost of inference in single-machine simulations. We measure prediction quality via the Brier Score ($=\frac{1}{N}\sum^{N}_{t=1}\sum^{R}_{i=1}(f_{ti}-o_{ti})^2$ where $N$ is the number of queries, $R$ is the number of target variables state combinations, while $f$ and $o$ are predicted and reference probabilities). We report the total processing time ratio between examined methods and the ground truth network for computation overhead. For communication we consider the count of factor values exchanged per query. We do not evaluate \augment{} time or communication as we expect it to be amortizable.

We test on public networks\footnote{https://www.bnlearn.com/bnrepository/} (\Cref{tab:net-stats}), consider two variable splitting methods, and multiple overlap variable ratios. \textbf{Related splits} assign to parties variables connected in the ground truth network. \textbf{Random splits} ensure parties have equal variable counts and share the same overlaps. Test sets contain 2000 queries, each with one random overlap variable as the target and 60\% of the others fixed as evidence.
Related splits attempt to simulate a realistic deployment in which the experts within clients have an incomplete but close-to-the-ground truth view of the interactions between their modeled variables (which also differ in number). Random splits aim for a worst-case scenario where the variables within clients (of roughly equal number) are not subgraphs of the complete network, generally giving more densely connected local networks and overlaps, as a worst-case scenario.

\subsection{Baselines}

As the closest prior work is strictly centralized, our baselines are:

\textbf{Centralized Combination (CC)} iteratively combines parties by the method of~\cite{combine_feng} and treats the network as an MRF if cycles form.

\textbf{Centralized Union (CU)}, the approach confidentially mimicked by \algo{}, is structurally based on the union of~\cite{combine_delsagrado}, combines parameters via geometric mean, and also applies the MRF principle.

\textbf{Decentralized Output Mean (DOM)} is a naïve approach that takes the geometric mean for each target variable's state probabilities over independently operating contained parties. It trades modeling long-range effects for confidentiality.

\textbf{\algoJ{}} is a degenerate \algo{} variant that stores the fully combined central CPDs for overlaps in one of the concerned parties, trading some safety for faster inference.

\subsection{\algo{} Performance Overview}

Here, we summarize the Brier score, computation time, and communication overhead of \algo{} under two splitting methods, different overlapping ratios (10\%, 30\%, and 50\%), and the party number (2, 4, and 8).  Due to the space limitations, we present the full results in the appendix and highlight in \Cref{fig:res} the performance trend via the representative case of four parties with a 30\% overlap ratio.

\textbf{Predictive Performance.} Regardless of split type, \algo{} predictions often outperform or match the classic CC and always beat the naïve DOM. \Cref{subfig:brier-related} gives results for related splits, \algo{} only scores worse than CC on the smallest network (0.015 versus 0.005) and gains a significant advantage over the larger ones (0.022 versus 0.009, 0.123 versus 0.114, and 0.006 versus 0.016). \algo{} has an advantage over DOM in all scenarios. Examining the random splits in \Cref{subfig:brier-random}, \algo{} still outscores CC in all but one of the smaller networks (0.031 vs 0.038) and maintains its lead over DOM in all but the smallest network, where the two tie. Since \algo{} yields the same predictive ability as its centralized counterpart CU, the differentiating points with CC are the structure combination policy and parameter combination operator. Over both splits, the pairing of the union selection and geometric mean aggregation in \algo{} fares better in most cases than the nondeterministic selection and superposition aggregation of CC, apart from some instances within small networks.
Finally, complete test results confirm the expectation that adding more parties tends to decrease performance while increasing overlaps has the opposite effect.

\textbf{Computation Overhead.} Regarding computation cost relative to centralized inference on the original network, average slowdowns are 1.63x for DOM, 1.15x for \algoJ{}, and 1.6x for \algo{}. Communication latency is unaccounted for as it can vary greatly based on the deployment. Still, we overestimate wall time by summing computation time across parties, as much processing would happen concurrently in reality. In the related splits from \Cref{subfig:time-related}, across the board, \algoJ{} is the fastest, but others follow closely. \algo{} is the second best in all except the largest networks, where it fares worse than DOM (1.17x vs 1.3x slower). The networks also examined for random splits in \Cref{subfig:time-random} show a similar trend, with somewhat higher general overhead, especially for \algo{}. Thus, as expected, \algoJ{} is faster than \algo{}, but both perform reasonably in most scenarios, even if the latter suffers more than the former as overlaps increase in complexity. The DOM implementation does seem to have a higher base overhead, but the gap to the other algorithms is often relatively contained. Further, complete results certify that adding parties improves speed while increasing overlaps decreases it. Regarding absolute total computation time, queries take the longest on the largest related splits dataset, with DOM averaging 2.9 ms/query, \algoJ{} 2.5 ms, and \algo{} 3.2 ms.

\textbf{Communication Cost.} Regarding the number of communicated CPD values per query, DOM averages merely 9, while \algoJ{} and \algo{} need orders of magnitude more at 387 and 1222, respectively. Since the number of communicated values depends on which party initiates a query, the reported figures include communication internal to the initiator to eliminate variability, tending to overestimate reality. The number of messages to complete a query is the same for all methods. Furthermore, the raw data transmitted in bytes remain in the low megabyte range for hundreds of thousands of values before compression. \Cref{subfig:comm-related} shows the mentioned discrepancy over related splits for all but the smallest network, in which the three methods are comparable. DOM merges complete party outputs and cannot propagate evidence between parties. Thus, it does not increase communication with the number of overlaps, and parties that do not contain any target variables send empty replies. The situation for random splits, illustrated in \Cref{subfig:comm-random}, is very similar, although the disadvantage of \algo{} over \algoJ{} widens considerably. As for adding parties and increasing overlaps, the complete appendix results attest that both increase communication for all methods.

\subsection{Party Weighting}

\begin{table}[tb]
\tabcolsep=0.1cm
\centering
\scalebox{0.775}{
\begin{tabular}{cl|cccc|cccc}
\multicolumn{2}{l|}{\#Parties} & \multicolumn{4}{c|}{2} & \multicolumn{4}{c}{4} \\
\multicolumn{2}{l|}{Vars in \textgreater{}1 Party} & \multicolumn{2}{c}{10\%} & \multicolumn{2}{c|}{30\%} & \multicolumn{2}{c}{10\%} & \multicolumn{2}{c}{30\%} \\
\multicolumn{2}{l|}{Method} & UW & W & UW & W & UW & W & UW & W \\ \hline
\multirow{4}{*}{\rotatebox[origin=c]{90}{Network}} & ASIA & 0.028 & \textbf{0.018} & 0.029 & \textbf{0.009} & 0.036 & \textbf{0.020} & 0.031 & \textbf{0.011} \\
 & CHILD & \textbf{0.067} & \textbf{0.067} & 0.088 & \textbf{0.047} & 0.084 & \textbf{0.067} & 0.097 & \textbf{0.048} \\
 & ALARM & 0.063 & \textbf{0.058} & 0.061 & \textbf{0.050} & 0.103 & \textbf{0.092} & 0.061 & \textbf{0.057} \\
 & INSURANCE & \textbf{0.079} & 0.104 & \textbf{0.051} & \textbf{0.051} & 0.112 & \textbf{0.088} & 0.135 & \textbf{0.098}
\end{tabular}
}
\caption{Unweighted \textbf{(UW)} \& Weighted \textbf{(W)} Brier scores for \algo{} relative to original network - random splits, imbalanced learning data (lower is better)}
\label{tab:brier-imba}
\end{table}

\Cref{tab:brier-imba} shows the weighted version of the proposed method having better predictive performance than the unweighted one in almost all scenarios with random splits. In weighting tests, we reduce the overall amount of data used for learning local BNs to ensure more variance and alternatingly assign parties a smaller or larger fraction of training data. Over the one scenario where the unweighted version performs better (0.079 versus 0.104), parties with lower data get overly punished for its perceived imprecision. Since each CPD has a single weight, all parents of a node within the party are still treated uniformly according to that value, even if there is a mismatch between the weight and the actual quality of the estimates. Similarly, if a party with lower overall confidence is the only one to model a highly-influential parent, its importance is slightly misrepresented in the final result. Nevertheless, although the strength of the effect has a considerably high variance, weighting has a clear positive overall impact in tested scenarios.

\subsection{Large Networks \& Many Parties}

\begin{table}[tb]
\tabcolsep=0.1cm
\centering
\scalebox{0.6}{
\begin{tabular}{l|ccc|ccc|ccc}
Networks/ & \multicolumn{3}{c|}{Brier Score} & \multicolumn{3}{c|}{Avg Comp Time Overhead} & \multicolumn{3}{c}{Avg \#Comm Values} \\
Parties & CC & DOM & CCBNet & DOM & CCBNetJ & CCBNet & DOM & CCBNetJ & CCBNet \\ \hline
ANDES/16 & 0.040 & 0.039 & \textbf{0.035} & 0.58 & 0.62 & 0.68 & 4 & 317 & 445 \\
PIGS/32 & 0.103 & 0.092 & \textbf{0.076} & 0.39 & 0.51 & 1.34 & 7 & 4202 & 37838 \\
LINK/64 & 0.144 & \textbf{0.125} & 0.126 & 0.25 & 0.35 & 0.39 & 6 & 2402 & 4549 \\
MUNIN2/128 & 0.016 & \textbf{0.015} & \textbf{0.015} & 0.20 & 0.41 & 0.67 & 11 & 88581 & 242788
\end{tabular}
}
\caption{Metrics for large networks \& many parties - related splits, 10\% of vars in \textgreater{}1 party (lower is better for all)}
\label{tab:large-nets}
\end{table}

Lastly, in \Cref{tab:large-nets}, our tests for challenging use cases with large networks (223-1003 features), many parties (16-28), and related variable splits, but lower overlaps reconfirm prediction/communication trends, yet computation improves over original networks. As previously, in larger networks, \algo{}'s predictions outperform CC, and communication size increases with parties and network size, averaging 71k values/request. Computation overhead is always \textless{}1 (i.e., a speedup) by 21\% on average, as the hardness of inference makes approximating a big network by splitting it into chunks faster, even before considering parallel party solving.

\section{Attacks on \algo{}}

\begin{figure}[tb]
\centering
\includegraphics[width=0.75\linewidth]{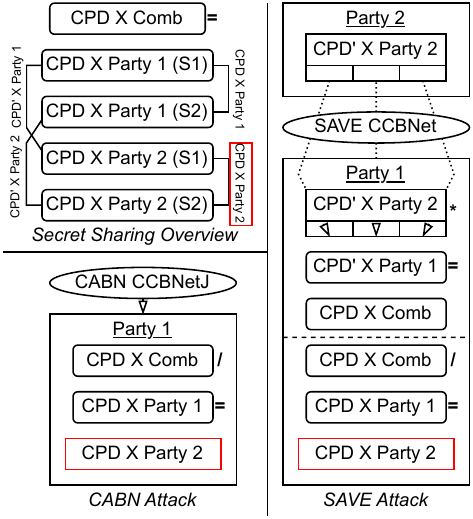}
\caption{Visualization of showcased \augment{} \& \infer{} attacks}
\hfill
\label{fig:attacks}
\end{figure}

To recap, a classic combination of related BNs, which encode confidential information into a single global model, has a very high risk of leaking information to all parties with direct access to it. As exemplified in the middle pane of \Cref{fig:comp}, even at a purely structural level, the centralized combination can contain much, if not all, of the local party information. Furthermore, at a parameter level, probability functions for any non-overlap nodes remain unmodified. Since BNs are human-readable, inspection can compromise sensitive information even before any inference.

We briefly review two attacks to reconstruct CPDs during \augment{} and \infer{}, respectively, along with their implications in \algo{} and \algoJ{}. The attacks do not bypass the obfuscation of unowned variable names and states but still expose potentially sensitive information via the recovered probability values. We successfully execute them, as visualized in \Cref{fig:attacks}, on a corner case of \textbf{two-party ASIA and CHILD network} instances.

\textbf{\augment{} Attack.} The first attack allows a party to reconstruct a peer's CPD for an overlap variable without inference, assuming no other parties are involved in the overlap, the attacker holds the combined CPD, and the protocol is \algoJ{}. The attacker starts by running \augment{} until it computes the combined CPD. Then, it removes its contribution from the geometric product that yielded the CPD, marginalizes any parents that should not be present in the victim's version, and normalizes to get the final result. With any secure computation scheme, if one of two input parties knows the output and the reversible operations applied, it can find the other input. For overlaps with three or more parties, the attacker can only reconstruct an aggregation of the other involved CPDs. Even when the attack is possible, name obfuscation still hides the real-world meaning of the variables previously unknown to the attacker. Should any party outside the overlap exist, in the assumed no collusion setting, having it compute and store the combined CPD instead gives a minimum-change fix. \algo{} is not vulnerable, as it does not join shares before applying inference operations.

\textbf{\infer{} Attack.} The second attack allows a party to reconstruct a peer's CPD for an overlap variable via inference, still assuming that the overlap contains no other parties but also applies to \algo{}. The attacker, who can be either of the two parties in the overlap, begins by querying for the overlap variable as the only target, specifying a state for each of its parents in the evidence. With all parents fixed, no other variables in the victim's network can affect the transmitted result. Thus, the attacker receives one row from the victim's share of the combined target CPD. Consequently, the attacker repeats the procedure for all other combinations of parent states, building up the victim's complete CPD share. It then takes the product of its share, and the one recovered from the other party to obtain the full combined CPD. Finally, like in the previous attack, it removes the contribution of its initial CPD (not the secret share) from the combination, marginalizes any parents not in the victim, and normalizes to find the result. Also similar to the previous attack, obfuscation limits the damage that can be done, while with the involvement of more than two parties, the attacker is able to compromise their shares, but not the contribution of each, before sharing the secrets. Thus, secret sharing avoids the attack for overlaps with three or more parties. As the attack requires a series of specific queries, redundant in most real applications, a simple defense has parties set a limit on the number of requests serviced that target a node with all parents in the evidence.

\section{Discussion}

\textbf{Framework Flexibility.} As outlined through the rest of the work, the framework naturally supports more variations than the presented \algo{} and \algoJ{}, allowing for a choice between increased confidentiality measures or other performance criteria. One choice entails replacing the HE with an SSS in \augment{} to swap computation by communication. Another is disallowing specific two-party overlap queries to avoid compromising network parameters. Of course, picking \algoJ{} for its speed or, conversely, \algo{} for its safety is perhaps the clearest such example.

For all variants above, changing \infer{} line \ref{inf-l4} to \texttt{VarElim(Q, \{\}, auxFacts, True)} so that inner variable elimination calls return the product of leftover factors instead of their set further increases security safeguards at the cost of overhead. Since the factors do not share variables, their product creates a comparatively much larger output to transmit. Nevertheless, it also makes it harder for the receiver to gain insight into the sender's independent variables.

\textbf{Framework Applicability.} Although our motivating use case comes from semiconductor manufacturing, we devise \algo{} to be as generally applicable as possible, even outside other manufacturing industries. Hence, we also base our evaluation on public data sets. Future improvements to further aid such goals could be confidentially harnessing available observed data samples within parties when updating overlap variable representation, or allowing additional types of workloads on the representation (e.g., approximate inference).

\section{Conclusion}

We propose \algo{} to address the issue of collaborative analysis for BNs in confidential (manufacturing) settings. The framework allows distributed analysis spanning involved models without revealing their contents. It has no model compatibility restrictions and allows unequally weighting parties. We extensively evaluate the method and a lower-overhead but less robust variant, \algoJ{}, against non-confidential central approaches and a naïve, distributed, confidential formulation. Results show \algo{} outperforming/matching the naïve/centralized approaches while maintaining reasonable computation time that decreases to 23\% of centralized formulations in large networks with many parties but produces much larger messages than minimum-interaction distributed alternatives, averaging 71k communicated values/request. Altogether, \algo{} offers centralized-like predictive performance in a distributed setting and ensures a base for protecting parties' private information.

\bibliography{refs-paper, refs-litho}

\pagebreak

\appendix

\section{Experiment Method \& Results Details}
\label{app:extra-eval}

\begin{table*}[tb!]
\tabcolsep=0.1cm
\centering
\scalebox{0.575}{
\begin{tabular}{cl|ccccccccc|cccccc|ccccccccc}
\multicolumn{2}{l|}{\#Parties} & \multicolumn{9}{c|}{2} & \multicolumn{6}{c|}{4} & \multicolumn{9}{c}{8} \\
\multicolumn{2}{l|}{Vars in \textgreater{}1 Party} & \multicolumn{3}{c}{10\%} & \multicolumn{3}{c}{30\%} & \multicolumn{3}{c|}{50\%} & \multicolumn{3}{c}{10\%} & \multicolumn{3}{c|}{50\%} & \multicolumn{3}{c}{10\%} & \multicolumn{3}{c}{30\%} & \multicolumn{3}{c}{50\%} \\
\multicolumn{2}{l|}{Method} & CC & DOM & CCBNet & CC & DOM & CCBNet & CC & DOM & CCBNet & CC & DOM & CCBNet & CC & DOM & CCBNet & CC & DOM & CCBNet & CC & DOM & CCBNet & CC & DOM & CCBNet \\ \hline
\multirow{5}{*}{\rotatebox[origin=c]{90}{Network}} & ASIA & 0.011 & 0.022 & \textbf{0.006} & 0.011 & 0.022 & \textbf{0.006} & \textbf{0} & 0.015 & 0.001 & \textbf{0.005} & 0.016 & 0.015 & \textbf{0.013} & 0.021 & 0.017 & \textbf{0.076} & \textbf{0.076} & \textbf{0.076} & \textbf{0.076} & \textbf{0.076} & \textbf{0.076} & 0.055 & \textbf{0.054} & \textbf{0.054} \\
 & CHILD & \textbf{0.002} & 0.034 & \textbf{0.002} & \textbf{0.001} & 0.02 & \textbf{0.001} & \textbf{0.001} & 0.02 & \textbf{0.001} & \textbf{0.08} & 0.106 & \textbf{0.08} & 0.012 & 0.033 & \textbf{0.011} & 0.165 & 0.173 & \textbf{0.164} & \textbf{0.071} & 0.09 & 0.072 & 0.044 & 0.046 & \textbf{0.023} \\
 & ALARM & \textbf{0.001} & 0.016 & 0.005 & \textbf{0.001} & 0.016 & 0.005 & \textbf{0.001} & 0.016 & 0.005 & \textbf{0.027} & 0.047 & 0.029 & 0.02 & 0.012 & \textbf{0.011} & 0.04 & 0.074 & \textbf{0.038} & 0.029 & 0.041 & \textbf{0.026} & 0.039 & 0.022 & \textbf{0.017} \\
 & INSURANCE & \textbf{0.007} & 0.022 & 0.009 & 0.015 & 0.017 & \textbf{0.011} & 0.013 & 0.01 & \textbf{0.008} & 0.111 & 0.117 & \textbf{0.108} & 0.086 & 0.037 & \textbf{0.032} & 0.154 & 0.166 & \textbf{0.153} & 0.189 & 0.135 & \textbf{0.131} & 0.091 & 0.085 & \textbf{0.077} \\
 & WIN95PTS & 0.015 & 0.007 & \textbf{0.001} & 0.01 & 0.007 & \textbf{0.005} & 0.01 & 0.007 & \textbf{0.005} & 0.019 & 0.011 & \textbf{0.005} & 0.01 & 0.007 & \textbf{0.004} & 0.026 & 0.016 & \textbf{0.011} & 0.038 & 0.016 & \textbf{0.011} & 0.016 & 0.013 & \textbf{0.006}
\end{tabular}
}
\caption{Brier score relative to original Network - related splits (lower is better)}
\label{tab:brier-related}
\end{table*}

\begin{table*}[tb!]
\tabcolsep=0.1cm
\centering
\scalebox{0.5}{
\begin{tabular}{cl|ccccccccc|cccccc|ccccccccc}
\multicolumn{2}{l|}{\#Parties} & \multicolumn{9}{c|}{2} & \multicolumn{6}{c|}{4} & \multicolumn{9}{c}{8} \\
\multicolumn{2}{l|}{Vars in \textgreater{}1 Party} & \multicolumn{3}{c}{10\%} & \multicolumn{3}{c}{30\%} & \multicolumn{3}{c|}{50\%} & \multicolumn{3}{c}{10\%} & \multicolumn{3}{c|}{50\%} & \multicolumn{3}{c}{10\%} & \multicolumn{3}{c}{30\%} & \multicolumn{3}{c}{50\%} \\
\multicolumn{2}{l|}{Method} & DOM & CCBNetJ & CCBNet & DOM & CCBNetJ & CCBNet & DOM & CCBNetJ & CCBNet & DOM & CCBNetJ & CCBNet & DOM & CCBNetJ & CCBNet & DOM & CCBNetJ & CCBNet & DOM & CCBNetJ & CCBNet & DOM & CCBNetJ & CCBNet \\ \hline
\multirow{5}{*}{\rotatebox[origin=c]{90}{Network}} & ASIA & 1.29 & 0.93 & 1.07 & 1.53 & 1.11 & 1.26 & 1.61 & 1.1 & 1.31 & 1.74 & 1.11 & 1.27 & 1.85 & 1.12 & 1.42 & 1.56 & 1.04 & 1.18 & 1.72 & 1.15 & 1.29 & 2.01 & 1.17 & 1.46 \\
 & CHILD & 1.16 & 1.01 & 1.07 & 1.39 & 1.02 & 1.19 & 1.4 & 1.03 & 1.2 & 1.2 & 1.0 & 1.11 & 1.72 & 1.15 & 1.54 & 1.26 & 1.01 & 1.07 & 1.5 & 1.03 & 1.25 & 2.06 & 1.13 & 1.81 \\
 & ALARM & 1.12 & 0.98 & 1.08 & 1.13 & 0.99 & 1.08 & 1.16 & 0.98 & 1.05 & 1.11 & 0.99 & 1.07 & 1.44 & 1.1 & 1.44 & 1.12 & 0.98 & 1.06 & 1.36 & 1.13 & 1.34 & 1.79 & 1.24 & 1.89 \\
 & INSURANCE & 1.18 & 0.99 & 1.09 & 1.36 & 1.07 & 1.31 & 1.56 & 1.11 & 1.49 & 1.18 & 1.06 & 1.19 & 1.66 & 1.2 & 1.65 & 1.2 & 1.02 & 1.17 & 1.34 & 1.05 & 1.31 & 1.8 & 1.13 & 1.73 \\
 & WIN95PTS & 1.03 & 0.92 & 1.01 & 1.13 & 0.96 & 1.11 & 1.13 & 0.97 & 1.13 & 0.93 & 0.89 & 0.97 & 1.28 & 1.09 & 1.49 & 0.89 & 0.85 & 0.91 & 1.18 & 1.04 & 1.37 & 1.45 & 1.08 & 1.73
\end{tabular}
}
\caption{Average computation time overhead relative to original network - related splits (lower is better)}
\label{tab:time-related}
\end{table*}

\begin{table*}[tb!]
\tabcolsep=0.1cm
\centering
\scalebox{0.5}{
\begin{tabular}{cl|ccccccccc|cccccc|ccccccccc}
\multicolumn{2}{l|}{\#Parties} & \multicolumn{9}{c|}{2} & \multicolumn{6}{c|}{4} & \multicolumn{9}{c}{8} \\
\multicolumn{2}{l|}{Vars in \textgreater{}1 Party} & \multicolumn{3}{c}{10\%} & \multicolumn{3}{c}{30\%} & \multicolumn{3}{c|}{50\%} & \multicolumn{3}{c}{10\%} & \multicolumn{3}{c|}{50\%} & \multicolumn{3}{c}{10\%} & \multicolumn{3}{c}{30\%} & \multicolumn{3}{c}{50\%} \\
\multicolumn{2}{l|}{Method} & DOM & CCBNetJ & CCBNet & DOM & CCBNetJ & CCBNet & DOM & CCBNetJ & CCBNet & DOM & CCBNetJ & CCBNet & DOM & CCBNetJ & CCBNet & DOM & CCBNetJ & CCBNet & DOM & CCBNetJ & CCBNet & DOM & CCBNetJ & CCBNet \\ \hline
\multirow{5}{*}{\rotatebox[origin=c]{90}{Network}} & ASIA & 4 & 16 & 22 & 4 & 16 & 22 & 4 & 16 & 24 & 4 & 12 & 17 & 4 & 14 & 23 & 4 & 11 & 15 & 4 & 11 & 15 & 4 & 12 & 20 \\
 & CHILD & 10 & 64 & 80 & 7 & 72 & 109 & 7 & 72 & 109 & 10 & 47 & 63 & 9 & 120 & 269 & 10 & 43 & 59 & 8 & 60 & 95 & 10 & 189 & 604 \\
 & ALARM & 5 & 65 & 93 & 5 & 65 & 93 & 5 & 65 & 93 & 6 & 100 & 146 & 7 & 484 & 1038 & 6 & 136 & 205 & 6 & 227 & 437 & 8 & 768 & 2070 \\
 & INSURANCE & 8 & 117 & 173 & 7 & 254 & 417 & 7 & 359 & 633 & 9 & 432 & 807 & 8 & 482 & 979 & 9 & 274 & 509 & 8 & 279 & 526 & 9 & 361 & 786 \\
 & WIN95PTS & 4 & 175 & 262 & 4 & 274 & 440 & 4 & 274 & 440 & 4 & 139 & 200 & 4 & 459 & 1000 & 4 & 115 & 158 & 5 & 344 & 1029 & 5 & 456 & 1588
\end{tabular}
}
\caption{Average \#communicated values rounded to nearest integer - related splits (lower is better)}
\label{tab:comm-related}
\end{table*}

\begin{table*}[tb!]
\tabcolsep=0.1cm
\centering
\scalebox{0.475}{
\begin{tabular}{ll|ccccccccc|ccccccccc|ccccccccc}
\multicolumn{2}{l|}{Metric} & \multicolumn{9}{c|}{Brier Score} & \multicolumn{9}{c|}{Average Computation Time Overhead} & \multicolumn{9}{c}{Average \#Communicated Values} \\
\multicolumn{2}{l|}{\#Parties} & \multicolumn{6}{c}{2} & \multicolumn{3}{c|}{4} & \multicolumn{6}{c}{2} & \multicolumn{3}{c|}{4} & \multicolumn{6}{c}{2} & \multicolumn{3}{c}{4} \\
\multicolumn{2}{l|}{Vars in \textgreater{}1 Party} & \multicolumn{3}{c}{10\%} & \multicolumn{3}{c}{30\%} & \multicolumn{3}{c|}{10\%} & \multicolumn{3}{c}{10\%} & \multicolumn{3}{c}{30\%} & \multicolumn{3}{c|}{10\%} & \multicolumn{3}{c}{10\%} & \multicolumn{3}{c}{30\%} & \multicolumn{3}{c}{10\%} \\
\multicolumn{2}{l|}{Method} & CC & DOM & CCBNet & CC & DOM & CCBNet & CC & DOM & CCBNet & DOM & CCBNetJ & CCBNet & DOM & CCBNetJ & CCBNet & DOM & CCBNetJ & CCBNet & DOM & CCBNetJ & CCBNet & DOM & CCBNetJ & CCBNet & DOM & CCBNetJ & CCBNet \\ \hline
\multirow{4}{*}{\rotatebox[origin=c]{90}{Network}} & ASIA & 0.021 & \textbf{0.012} & \textbf{0.012} & 0.011 & \textbf{0.006} & \textbf{0.006} & 0.029 & \textbf{0.023} & \textbf{0.023} & 1.18 & 0.92 & 1.0 & 1.49 & 1.08 & 1.24 & 1.73 & 1.19 & 1.47 & 4 & 11 & 15 & 4 & 11 & 16 & 8 & 16 & 38 \\
 & CHILD & 0.031 & \textbf{0.021} & 0.023 & 0.027 & 0.02 & \textbf{0.018} & \textbf{0.029} & 0.047 & 0.031 & 1.23 & 1.06 & 1.13 & 1.36 & 1.08 & 1.33 & 1.37 & 1.07 & 1.34 & 6 & 97 & 146 & 7 & 153 & 254 & 12 & 80 & 202 \\
 & ALARM & \textbf{0.009} & 0.013 & 0.011 & 0.006 & 0.008 & \textbf{0.005} & 0.025 & 0.043 & \textbf{0.022} & 1.13 & 0.98 & 1.06 & 1.31 & 1.07 & 1.29 & 1.27 & 1.15 & 1.69 & 5 & 189 & 304 & 6 & 423 & 691 & 11 & 714 & 2599 \\
 & INSURANCE & 0.05 & 0.031 & \textbf{0.028} & 0.039 & 0.025 & \textbf{0.024} & 0.072 & 0.068 & \textbf{0.052} & 1.15 & 1.04 & 1.18 & 1.33 & 1.14 & 1.4 & 1.33 & 1.21 & 1.79 & 6 & 195 & 315 & 6 & 336 & 561 & 12 & 686 & 2464
\end{tabular}
}
\caption{Results for random splits (lower is better for all)}
\label{tab:rand-nets}
\end{table*}

\begin{algorithm}[tb]
\textbf{Input}: bnDAG, nrSplits, nrOverlaps, randGen\\
\textbf{Output}: Splits
\begin{algorithmic}[1]
    \STATE dfsTree $\leftarrow$ \textit{DFSTree}(bnDAG)
    \STATE splits $\leftarrow$ \textit{GreedyModularityComms}(DFSTree, nrSplits)\\\cite{communities_clauset}
    \STATE shuffledEdges $\leftarrow$ randGen.\textit{shuffle}(bnDAG.edges)
    \STATE initNodeSplit $\leftarrow$ \{\}
    \FOR{splitNr $\leftarrow$ 1 $\dots$ nrSplits}
        \FOR{node $\leftarrow$ splits[nrSplit]}
            \STATE initNodeSplit[node] $\leftarrow$ splitNr
        \ENDFOR
    \ENDFOR
    \STATE ovNodes $\leftarrow$ \{ \}
    \STATE connSplits $\leftarrow$ \{ \}
    \STATE extraEdges $\leftarrow$ [ ]
    \FOR{nodeO, nodeI $\leftarrow$ shuffledEdges}
        \IF{$|$ovNodes$|$ $\geq$ nrOverlaps}
            \STATE break
        \ENDIF
        \STATE es $\leftarrow$ \{initNodeSplit[nodeO], initNodeSplit[nodeI]\}
        \IF{$|$es$|$ == 1}
            \STATE continue
        \ENDIF
        \IF{$|$connSplits$|$ \textless{} nrSplits AND es $\subseteq$ connSplits}
            \STATE extraEdges += (nodeO, nodeI)
        \ELSE
            \STATE ovNodes $\cup$= \{nodeO, nodeI\}
            \STATE connSplits $\cup$= es
            \STATE splits[initNodeSplit[nodeO]] $\cup$= \{nodeO\}
            \STATE splits[initNodeSplit[nodeI]] $\cup$= \{nodeI\}
        \ENDIF
    \ENDFOR
    \FOR{nodeO, nodeI $\leftarrow$ extraEdges}
        \IF{$|$ovNodes$|$ $\geq$ nrOverlaps}
            \STATE break
        \ENDIF
        \STATE ovNodes $\cup$= \{nodeO, nodeI\}
        \STATE splits[initNodeSplit[nodeO]] $\cup$= \{nodeO\}
        \STATE splits[initNodeSplit[nodeI]] $\cup$= \{nodeI\}
    \ENDFOR
    \RETURN splits
\end{algorithmic}
\caption{Related Split}
\label{alg:related-split}
\end{algorithm}

\begin{algorithm}[tb]
\textbf{Input}: bnDAG, nrSplits, nrOverlaps, randGen\\
\textbf{Output}: Communities
\begin{algorithmic}[1]
    \STATE shuffledNodes $\leftarrow$ randGen.\textit{shuffle}(bnDAG.nodes)
    \STATE ovs $\leftarrow$ randGen.\textit{sample}(shuffledNodes, nrOverlaps)
    \STATE splits $\leftarrow$ \textit{SplitEqualParts}(shuffledNodes, nrSplits)
    \FOR{split $\leftarrow$ splits}
        \STATE split $\cup$= ovs
    \ENDFOR
    \RETURN splits
\end{algorithmic}
\caption{Random Split}
\label{alg:random-split}
\end{algorithm}

We detail the procedures used for related and random splits of ground truth variables amongst parties, used throughout all experiments, in \Cref{alg:related-split} and \Cref{alg:random-split}, respectively. \Cref{tab:brier-related}, \Cref{tab:time-related}, and \Cref{tab:comm-related}. Regarding additional experiment results with related splits, \Cref{tab:brier-related} covers predictive performance, \Cref{tab:time-related} covers computation overhead, and \Cref{tab:comm-related} covers communication cost. In a few networks under related splits, adjacent overlap figures (e.g., 30\% and 50\%) have the same splits and inherently score, either because of the small number of nodes (ASIA) or because all possible overlaps for the given topology are formed (CHILD, ALARM). \Cref{tab:rand-nets} shows extra results for all three metrics under random splits.
For choosing the order of variable elimination during inference, we use a min weight heuristic, which greedily chooses a variable such that the product of factors containing it has the smallest size.

We run each experiment once, with a fixed seed determining the randomness for sampling data instances from the reference network and splitting it into overlapping variables sets of parties. We utilize an Intel(R) Core(TM) i9-12900KF CPU @ 3.20GHz (note that inference in the single-machine simulation is single-threaded), with 120 GB of RAM, and Ubuntu 20.04.6 LTS as the OS. We implement the codebase in Python 3.10, using pgmpy 0.1.24\footnote{https://github.com/pgmpy/pgmpy} as the backbone for BNs in our framework, tenseal 0.3.14\footnote{https://github.com/OpenMined/TenSEAL} for homomorphic encryption, and openmined.psi 2.0.2\footnote{https://github.com/OpenMined/PSI} for private set intersection. We provide the code as supplementary material.

\section{Multiplication-based Secret Sharing}
\label{app:multip-sss}

\begin{definition}[Group]
A group~\cite{sss_kales} is a set $G (\neq \emptyset)$ and operation $\bullet: G \times G \longrightarrow G$ such that:
\begin{enumerate}
    \item Asociativity: $a \bullet (b \bullet c) = (a \bullet b) \bullet c, \forall a, b, c \in G$
    \item Neutral element: $\exists! e \in G$ such that $ e \bullet a = a \bullet e = a, \forall a \in G$
    \item Inverse element:$a \bullet a' = a' \bullet a = e, \forall a \in G, \exists!a'$, where $e$ is the neutral element
\end{enumerate}
\label{def:group}
\end{definition}

\begin{definition}[$\mathbb{Z}_n$]
$\mathbb{Z}_n$ is a group under $\{0, 1, \dots, n-1\}$ and addition modulo (\%) $n$.
\label{def:zn}
\end{definition}

\begin{definition}[$\mathbb{Z}_p^*$]
$\mathbb{Z}_p^*$, for $p$ prime, is a group under $\{1, 2, \dots, p-1\}$ and multiplication modulo (\%) $p$.
\label{def:zpstar}
\end{definition}

\begin{proposition}
The inverse $x'$ of $x \in \mathbb{Z}_p^*$ is $x^{p-2} \% p$.
\label{prop:inv}
\end{proposition}

\begin{figure}[tb]
\centering
\includegraphics[width=1.0\linewidth]{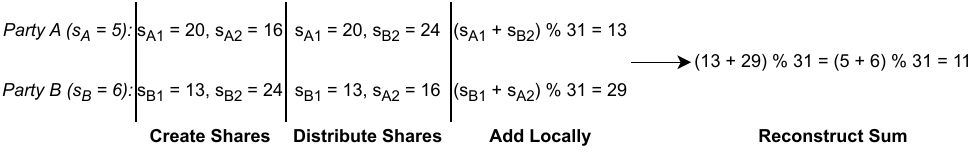}
\caption{Example of computing the sum of two parties' additively secret shared variables in $\mathbb{Z}_{31}$}
\hfill
\label{fig:add-sss}
\end{figure}

\begin{figure}[tb]
\centering
\includegraphics[width=1.0\linewidth]{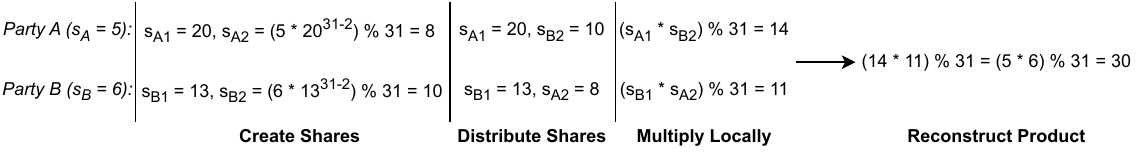}
\caption{Example of computing the product of two parties' multiplication-based secret shared variables in $\mathbb{Z}_{31}^*$}
\hfill
\label{fig:multip-sss}
\end{figure}

Secret sharing schemes are a family of methods that allow the distribution of a secret value among a group of parties by assigning each a share that does not yield any information about the secret but can, when pooled with enough others, reveal it~\cite{smpc_cramer}. The following details the classic additive scheme and the multiplication-based version used for \algo{}, relating them to each other.

Not leaking any information about a secret with theoretical guarantees for an adversary of unbounded computation requires sampling it from a uniform distribution, which is impossible for any infinite set, like $\mathbb{Z}$ or any subset of $\mathbb{R}$. Thus, most schemes perform their computation within (derivatives of) finite integer groups (\Cref{def:group}) large enough to contain all possibly required values for computations on the secrets. The previously mentioned fixed-precision encoding is an usual method for incorporating floating-point values. Multiplication in linear schemes requires additional interaction between the parties. Usually, it involves the help of a secretly sharing an additional set of values $a, b, c$, called Beaver triples~\cite{triples_beaver}, obeying $a * b = c$, with $a$ and $b$ chosen arbitrarily. Protocols exist for parties to generate triples amongst themselves securely, but a trusted third party can simplify the procedure.

Additive~\cite{sss_kales} secret sharing is a popular scheme in literature, which is defined on $\mathbb{Z}_n$ (\Cref{def:zn}), with information-theoretic security for up to $n-1$ passively corrupt parties. \Cref{fig:add-sss} exemplifies adding two parties' secret values. Shamir's secret sharing has a configurable $(t, n)$ threshold scheme, which produces secret shares based on a polynomial whose constant coefficient is the secret, and all others are random. It has information-theoretic security for up to $\lfloor n/2 \rfloor$ passive corrupt parties and $\lfloor n/3 \rfloor$ active ones. It uses Beaver triples to allow multiplication.

The multiplication-based scheme utilized in \algo{} is similar to the additive one but becomes non-linear by swapping efficient share addition for multiplication. As it works under $\mathbb{Z}_p^*$ (\Cref{def:zpstar}), not $\mathbb{Z}_n$, instead of agreeing on a large enough $n$, parties agree on a large enough prime $p$ to instantiate the group. As in the additive case, to split a secret $s$ into $k$ shares, parties get shares $s_1$ through $s_{k-1}$ by uniformly sampling the group's set of values and fix $s_k = s \bullet (s_1 \bullet ... \bullet s_{k-1})'$. From \Cref{prop:inv}, it follows that $s_k = s * (s_1 * ... * s_{k-1})^{p-2}$, where all multiplications are modulo $p$. Modular exponentiation can be efficiently computed even for large exponents. Reconstruction still happens by applying the group operator to all shares. Note that, although $0 \notin \mathbb{Z}_p^*$, assuming that the party holding the secret keeps one of the shares, it can set its share to $0$, and sample $\mathbb{Z}_p^*$ for the remaining ones. \Cref{fig:multip-sss} gives a small example of multiplying two parties' secret values.

\end{document}